\documentclass[10pt,twocolumn,letterpaper]{article}

\usepackage{cvpr}
\usepackage{times}
\usepackage{epsfig}
\usepackage{graphicx}
\usepackage{amsmath}
\usepackage{amssymb}
\usepackage{algorithm}
\usepackage{paralist}

\DeclareMathOperator*{\argmax}{arg\,max}
\usepackage{algorithmic}

\usepackage[pagebackref=true,breaklinks=true,letterpaper=true,colorlinks,bookmarks=false]{hyperref}

\cvprfinalcopy 

\def\cvprPaperID{4023} 

\ifcvprfinal\fi
\begin{document}

\newcommand{\textred}[1]{\textcolor{red}{#1}}
\ifx\noeditingmarks\undefined
   \newcommand{\pgwrapper}[2]{\textred{#1: #2}}
\else
   \newcommand{\pgwrapper}[2]{}
\fi
\newcommand{\hb}[1]{\pgwrapper{HB}{#1}}
\newcommand{\srm}[1]{\pgwrapper{SAM}{#1}}
\newcommand{\ma}[1]{\pgwrapper{MA}{#1}}
\newcommand{\sanjay}[1]{\pgwrapper{SC}{#1}}
\newcommand{\sofa}[1]{\pgwrapper{SA}{#1}}
\title{RoadTracer: Automatic Extraction of Road Networks from Aerial Images}

\author{Favyen Bastani\textsuperscript{1}, Songtao He\textsuperscript{1}, Sofiane Abbar\textsuperscript{2}, Mohammad Alizadeh\textsuperscript{1}, Hari Balakrishnan\textsuperscript{1},\\Sanjay Chawla\textsuperscript{2}, Sam Madden\textsuperscript{1}, David DeWitt\textsuperscript{1} \\
\textsuperscript{1}MIT CSAIL, \textsuperscript{2}Qatar Computing Research Institute, HBKU \\
 \textsuperscript{1}\{fbastani,songtao,alizadeh,hari,madden,dewitt\}@csail.mit.edu, \textsuperscript{2}\{sabbar,schawla\}@hbku.edu.qa
}


\maketitle

\begin{abstract}

Mapping road networks is currently both expensive and labor-intensive. High-resolution aerial imagery provides a promising avenue to automatically infer a road network. Prior work uses convolutional neural networks (CNNs) to detect which pixels belong to a road (segmentation), and then uses complex post-processing heuristics to infer graph connectivity. We show that these segmentation methods have high error rates because noisy CNN outputs are difficult to correct. We propose RoadTracer, a new method to automatically construct accurate road network maps from aerial images. RoadTracer uses an iterative search process guided by a CNN-based decision function to derive the road network graph directly from the output of the CNN. We compare our approach with a segmentation method on fifteen cities, and find that at a 5\% error rate, RoadTracer correctly captures 45\% more junctions across these cities.

\end{abstract}

\vspace{-0.2in}
\section{Introduction} \label{sec:introduction}

Creating and updating road maps is a tedious, expensive, and often manual process today~\cite{wired-gmaps}. Accurate and up-to-date maps are especially important given the popularity of location-based mobile services and the impending arrival of autonomous vehicles. Several companies are investing hundreds of millions of dollars on mapping the world, but despite this investment, error rates are not small in practice, with map providers receiving many tens of thousands of error reports per day.\footnote{See, \eg, \url{https://productforums.google.com/forum/\#!topic/maps/dwtCso9owlU} for an example of a city (Doha, Qatar) where maps have been missing entire subdivisions for years.} In fact, even obtaining ``ground truth'' maps in well-traveled areas may be difficult; recent work~\cite{mattyus2017deeproadmapper} reported that the discrepancy between OpenStreetMap (OSM) and the TorontoCity dataset was 14\% (the recall according to a certain metric for OSM was 0.86).

\begin{figure}[t]
\begin{center}
	\includegraphics[width=\linewidth]{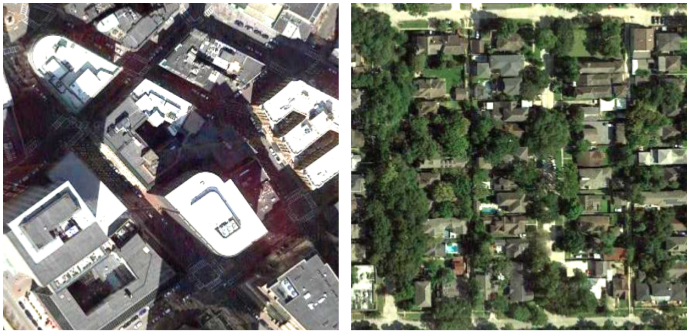}
\end{center}
	\caption{Occlusions by trees, buildings, and shadows make it hard even for humans to infer road connectivity from images.}
\label{fig:occluded_roads}
\vspace{-.2in}
\end{figure}


Aerial imagery provides a promising avenue to automatically infer the road network graph. In practice, however, extracting maps from aerial images is difficult because of occlusion by trees, buildings, and shadows (see Figure \ref{fig:occluded_roads}). Prior approaches do not handle these problems well. Almost universally, they begin by segmenting the image, classifying each pixel in the input as either road or non-road~\cite{cheng2017automatic,mattyus2017deeproadmapper}. They then implement a complex post-processing pipeline to interpret the segmentation output and extract topological structure to construct a map. As we will demonstrate, noise frequently appears in the segmentation output, making it hard for the post-processing steps to produce an accurate result.

The fundamental problem with a segmentation-based approach is that the CNN is trained only to provide local information about the presence of roads. Key decisions on how road segments are inter-connected to each other are delegated to an error-prone post-processing stage that relies on heuristics instead of machine learning or principled algorithms. Rather than rely on an intermediate image representation, we seek an approach that produces the road network directly from the CNN. However, it is not obvious how to train a CNN to learn to produce a graph from images.

We propose RoadTracer, an approach that uses an {\em iterative graph construction} process for extracting graph structures from images. Our approach constructs the road network by adding individual road segments one at a time, using a novel CNN architecture to decide on the next segment to add given as input the portion of the network constructed so far. In this way, we eliminate the intermediate image representation of the road network, and avoid the need for extensive post-processing that limits the accuracy of prior methods.

Training the CNN decision function is challenging because the input to the CNN at each step of the search depends on the partial road network generated using the CNN up to that step. We find that standard approaches that use a static set of labeled training examples are inadequate. Instead, we develop a dynamic labeling approach to produce training examples on the fly as the CNN evolves during training. This procedure resembles reinforcement learning, but we use it in an efficient supervised training procedure.

We evaluate our approach using aerial images covering 24 square km areas of 15 cities, after training the model on 25 other cities. We make our code and a demonstration of RoadTracer in action available at \url{https://roadmaps.csail.mit.edu/roadtracer}. We implement two baselines, DeepRoadMapper~\cite{mattyus2017deeproadmapper} and our own segmentation approach. Across the 15 cities, our main experimental finding is that, at a 5\% average error rate on a junction-by-junction matching metric, RoadTracer correctly captures 45\% more junctions than our segmentation approach (0.58 vs 0.40). DeepRoadMapper fails to produce maps with better than a 19\% average error rate. Because accurately capturing the local topology around junctions is crucial for applications like navigation, these results suggest that RoadTracer is an important step forward in fully automating map construction from aerial images.

\section{Related Work} \label{sec:related}

Classifying pixels in an aerial image as ``road'' or ``non-road'' is a well-studied problem, with solutions generally using probabilistic models. Barzobar \etal build geometric-probabilistic models of road images based on assumptions about local road-like features, such as road geometry and color intensity, and draw inferences with MAP estimation~\cite{barzohar1996automatic}. Wegner \etal use higher-order conditional random fields (CRFs) to model the structures of the road network by first segmenting aerial images into superpixels, and then adding paths to connect these superpixels~\cite{wegner2015road}. More recently, CNNs have been applied to road segmentation~\cite{mnih2010learning,costea2016aerial}. However, the output of road segmentation, consisting of a probability of each pixel being part of a road, cannot be directly used as a road network graph.


To extract a road network graph from the segmentation output, Cheng \etal apply binary thresholding and morphological thinning to produce single-pixel-width road centerlines~\cite{cheng2017automatic}. A graph can then be obtained by tracing these centerlines. M{\'a}ttyus \etal propose a similar approach called DeepRoadMapper, but add post-processing stages to enhance the graph by reasoning about missing connections and applying heuristics~\cite{mattyus2017deeproadmapper}. This solution yields promising results when the road segmentation has modest error. However, as we will show in Section \ref{sec:segmentation}, heuristics do not perform well when there is uncertainty in segmentation, which can arise due to occlusion, ambiguous topology, or complex topology such as parallel roads and multi-layer roads.


Rather than extract the road graph from the result of segmentation, some solutions directly extract a graph from images. Hinz \etal produce a road network using a complex road model that is built using detailed knowledge about roads and their context, such as nearby buildings and vehicles~\cite{hinz2003automatic}. Hu \etal introduce road footprints, which are detected based on shape classification of the homogeneous region around a pixel~\cite{hu2007road}. A road tree is then grown by tracking these road footprints. Although these approaches do not use segmentation, they involve numerous heuristics and assumptions that resemble those in the post-processing pipeline of segmentation-based approaches, and thus are susceptible to similar issues.


Inferring road maps from GPS trajectories has also been studied~\cite{biagioni2012map,stanojevic2017kharita,cobweb}. However, collecting enough GPS data that can cover the entire map in both space and time is challenging, especially when the region of the map is large and far from the city core. Nevertheless, GPS trajectories may be useful to improve accuracy in areas where roads are not visible from the imagery, to infer road connectivity at complex interchanges where roads are layered, and to enable more frequent map updates.




\section{Automatic Map Inference} \label{sec:method}

The goal of automatic map inference is to produce a road network map, i.e., a graph where vertices are annotated with spatial coordinates (latitude and longitude), and edges correspond to straight-line road segments. Vertices with three or more incident edges correspond to road junctions (e.g. intersections  or forks). Like prior methods, we focus on inferring undirected road network maps, since the directionality of roads is generally not visible from aerial imagery.

In Section \ref{sec:segmentation}, we present an overview of segmentation-based map-inference methods used by current state-of-the-art techniques~\cite{cheng2017automatic,mattyus2017deeproadmapper} to construct a road network map from aerial images. We describe problems in the maps inferred by the segmentation approach to motivate our alternative solution. Then, in Section \ref{sec:iterative}, we introduce our iterative map construction method. In Section \ref{sec:training}, we discuss the procedure used to train the CNN used in our solution.

\subsection{Prior Work: Segmentation Approaches} \label{sec:segmentation}


Segmentation-based approaches have two steps. First, each pixel is labeled as either ``road'' or ``non-road''. Then, a post-processing step applies a set of heuristics to convert the segmentation output to a road network graph.

State-of-the-art techniques share a similar post-processing pipeline to extract an initial graph from the segmentation output. The segmentation output is first thresholded to obtain a binary mask. Then, they apply morphological thinning~\cite{thinning} to produce a mask where roads are represented as one-pixel-wide centerlines. This mask is interpreted as a graph, where set pixels are vertices and edges connect adjacent set pixels. The graph is simplified with the Douglas-Peucker method~\cite{douglas1973algorithms}.



\begin{figure}[t]
\begin{center}
	\includegraphics[width=\linewidth]{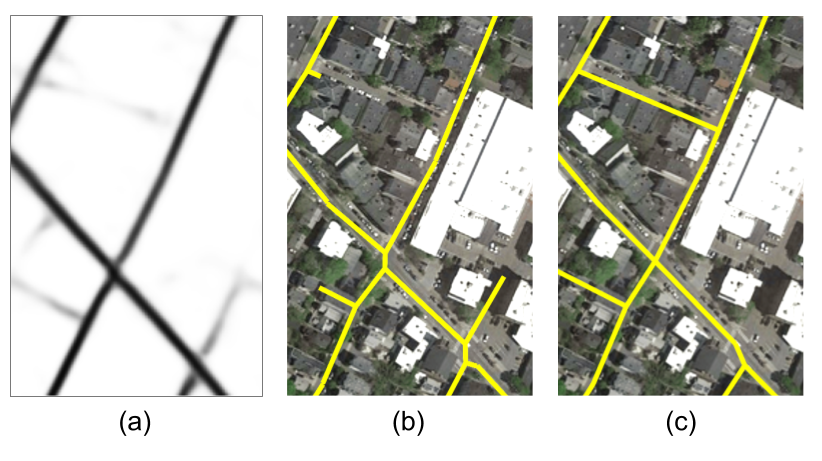}
\end{center}
	\caption{Stages of segmentation post-processing. (a) shows the segmentation output. In (b), a graph is extracted through morphological thinning~\cite{thinning} and the Douglas-Peucker method~\cite{douglas1973algorithms}. Refinement heuristics are then applied to remove basic types of noise, yielding the graph in (c).}
\label{fig:segmentation_cleaning}
\end{figure}

Because the CNN is trained with a loss function evaluated independently on each pixel, it will yield a noisy output in regions where it is unsure about the presence of a road. As shown in Figure \ref{fig:segmentation_cleaning}(a) and (b), noise in the segmentation output will be reflected in the extracted graph. Thus, several methods have been proposed to refine the initial extracted graph. Figure \ref{fig:segmentation_cleaning}(c) shows the graph after applying three refinement heuristics: pruning short dangling segments, extending dead-end segments, and merging nearby junctions.



\begin{figure}[t]
\begin{center}
	\includegraphics[width=\linewidth]{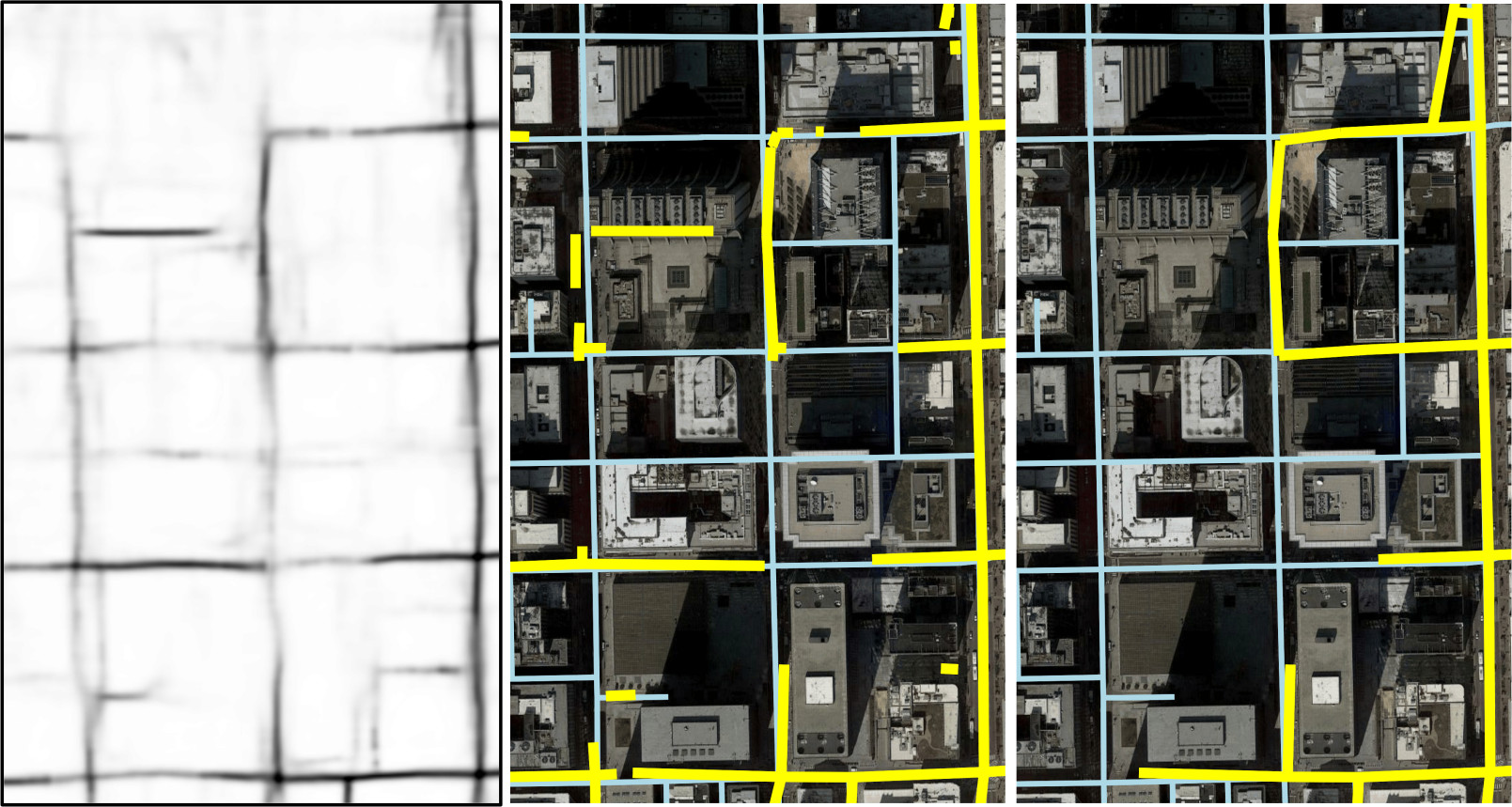}
\end{center}
	\caption{An example where noise in the segmentation output (left) is too extensive for refinement heuristics to correct. We show the graph after refinement on the right. Here, we overlay the inferred graph (yellow) over ground truth from OSM (blue).}
\label{fig:segmentation_noise}
\end{figure}

Although refinement is sufficient to remove basic types of noise, as in Figure \ref{fig:segmentation_cleaning}, we find that many forms of noise are too extensive to compensate for. In Figure \ref{fig:segmentation_noise}, we show an example where the segmentation output contains many gaps, leading to a disconnected graph with poor coverage. Given this segmentation output, even a human would find it difficult to accurately map the road network. Because the CNN is trained only to classify individual pixels in an image as roads, it leaves us with an untenable jigsaw puzzle of deciding which pixels form the road centerlines, and where these centerlines should be connected.



These findings convinced us that we need a different approach that can produce a road network directly, {\em without} going through the noisy intermediate image representation of the road network. We propose an iterative graph construction architecture to do this. By breaking down the mapping process into a series of steps that build a road network graph iteratively, we will show that we can derive a road network from the CNN, thereby eliminating the requirement of a complex post-processing pipeline and yielding more accurate maps.

\subsection{RoadTracer: Iterative Graph Construction} \label{sec:iterative}

In contrast to the segmentation approach, our approach consists of a search algorithm, guided by a decision function implemented via a CNN, to compute the graph iteratively. The search walks along roads starting from a single location known to be on the road network. Vertices and edges are added in the path that the search follows. The decision function is invoked at each step to determine the best action to take: either add an edge to the road network, or step back to the previous vertex in the search tree. Algorithm~\ref{alg:search} shows the pseudocode for the search procedure.


\begin{algorithm}[t]
\caption{Iterative Graph Construction}
\label{alg:search}
\begin{algorithmic}
    \REQUIRE A starting location $v_0$ and the bounding box $B$
    \STATE{initialize graph $G$ and vertex stack $S$ with $v_0$}
    \WHILE{$S$ is not empty}
        \STATE{$action, \alpha := \text{decision\_func}(G, S_{\text{top}}, \mathit{Image}$)}
        \STATE{$u := S_{\text{top}} + (D \cos{\alpha}, D \sin{\alpha})$}
        \IF{$\mathit{action} = \text{stop}$ or $u$ is outside $B$}
            \STATE{pop $S_{\text{top}}$ from $S$}
        \ELSE
            \STATE{add vertex $u$ to $G$}
            \STATE{add an edge $(S_{\text{top}}, u)$ to $G$}
            \STATE{push $u$ onto $S$}
        \ENDIF
    \ENDWHILE
\end{algorithmic}
\end{algorithm}

\smallskip
\noindent{\bf Search algorithm.}
We input a region $(v_0, B)$, where $v_0$ is the known starting location, and $B$ is a bounding box defining the area in which we want to infer the road network. The search algorithm maintains a graph $G$ and a stack of vertices $S$ that both initially contain only the single vertex $v_0$. $S_{\text{top}}$, the vertex at the top of $S$, represents the current location of the search.

At each step, the decision function is presented with $G$, $S_{\text{top}}$, and an aerial image centered at $S_{top}$'s location. It can decide either to walk a fixed distance $D$ (we use $D = 12$ meters) forward from $S_{\text{top}}$ along a certain direction, or to {\em stop} and return to the vertex preceding $S_{\text{top}}$ in $S$. When walking, the decision function selects the direction from a set of $a$ angles that are uniformly distributed in $[0, 2\pi)$. Then, the search algorithm adds a vertex $u$ at the new location (i.e., $D$ away from $S_{\text{top}}$ along the selected angle), along with an edge $(S_{\text{top}}, u)$, and pushes $u$ onto $S$ (in effect moving the search to $u$).



If the decision process decides to ``stop'' at any step, we pop $S_{\text{top}}$ from $S$. Stopping indicates that there are no more unexplored roads (directions) adjacent to $S_{\text{top}}$. Note that because only new vertices are ever pushed onto $S$, a ``stop'' means that the search will never visit the vertex $S_{\text{top}}$ again.

\begin{figure}[t]
\begin{center}
	\includegraphics[width=\linewidth]{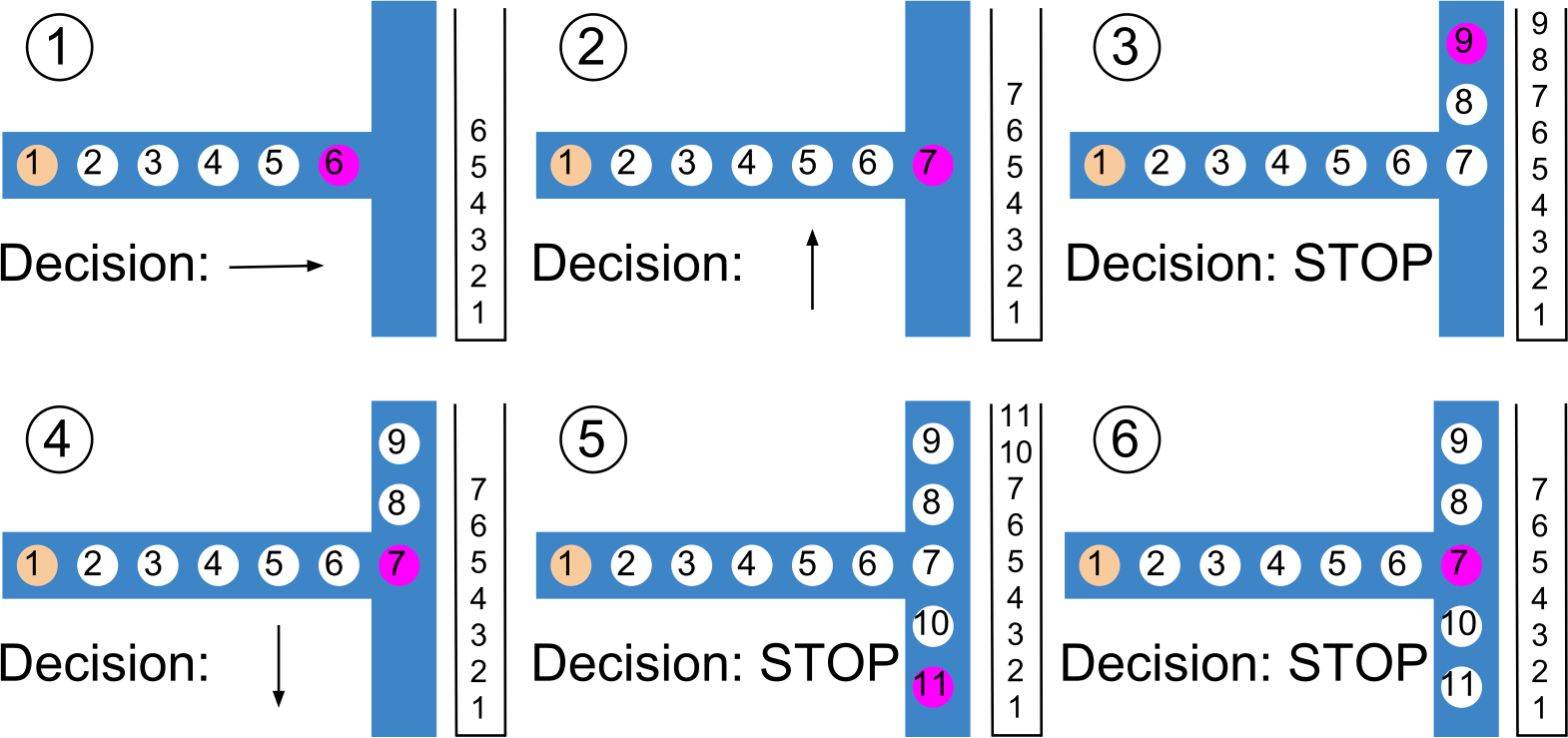}
\end{center}
	\caption{Exploring a T intersection in the search process. The blue path represents the position of the road in the satellite imagery. Circles are vertices in $G$, with $S_{\text{top}}$ in purple and $v_0$ in orange. Here, the decision function makes correct decisions on each step.}
\label{fig:actor_t_intersection}
\end{figure}


Figure~\ref{fig:actor_t_intersection} shows an example of how the search proceeds at an intersection. When we reach the intersection, we first follow the upper branch, and once we reach the end of this branch, the decision function selects the ``stop'' action. Then, the search returns to each vertex previously explored along the left branch. Because there are no other roads adjacent to the upper branch, the decision function continues to select the stop action until we come back to the intersection. At the intersection, the decision function leads the search down the lower branch. Once we reach the end of this branch, the decision function repeatedly selects the stop action until we come back to $v_0$ and $S$ becomes empty. When $S$ is empty, the construction of the road network is complete.

Since road networks consist of cycles, it is also possible that we will turn back on an earlier explored path. The search algorithm includes a simple merging step to handle this: when processing a walk action, if $u$ is within distance $3D$ of a vertex $v \in G$, but the shortest distance in $G$ from $S_{\text{top}}$ to $v$ is at least $6D$, then we add an edge $(u, v)$ and don't push $u$ onto $S$. This heuristic prevents small loops from being created, e.g. if a road forks into two at a small angle.


Lastly, we may walk out of our bounding box $B$. To avoid this, when processing a walk action, if $u$ is not contained in $B$, then we treat it as a stop action.

\smallskip
\noindent{\bf CNN decision function.}
A crucial component of our algorithm is the decision function, which we implement with a CNN. The input layer consists of a $d \times d$ window centered on $S_{\text{top}}$. This window has four channels.  The first three channels are the RGB values of the $d \times d$ portion of aerial imagery around $S_{\text{top}}$. The fourth channel is the graph constructed so far, $G$. We render $G$ by drawing anti-aliased lines along the edges of $G$ that fall inside the window. Including $G$ in the input to the CNN is a noteworthy aspect of our method. First, this allows the CNN to understand which roads in the aerial imagery have been explored earlier in the search, in effect moving the problem of excluding these roads from post-processing to the CNN. Second, it provides the CNN with useful context; \eg, when encountering a portion of aerial imagery occluded by a tall building, the CNN can use the presence or absence of edges on either side of the building to help determine whether the building occludes a road.


The output layer consists of two components: an action component that decides between walking and stopping, and and an angle component that decides which angle to walk in. The action component is a softmax layer with 2 outputs, $O_{action} = \langle o_{walk}, o_{stop} \rangle$. The angle component is a sigmoid layer with $a$ neurons, $O_{angle} = \langle o_1, \ldots, o_a \rangle$. Each $o_i$ corresponds to an angle to walk in. We use a threshold to decide between walking and stopping. If $o_{walk} \geq T$, then walk in the angle corresponding to $\argmax_i(o_i)$. Otherwise, stop.







We noted earlier that our solution does not require complex post-processing heuristics, unlike segmentation-based methods where CNN outputs are noisy. The only post-processing required in our decision function is to check a threshold on the CNN outputs and select the maximum index of the output vector. Thus, our method enables the CNN to directly produce a road network graph.
\section{Iterative Graph Construction CNN Training} \label{sec:training}

We now discuss the training procedure for the decision function. We assume we have a ground truth map $G^{*}$ (\eg, from OpenStreetMap). Training the CNN is non-trivial: the CNN takes as input a partial graph $G$ (generated by the search algorithm) and outputs the desirability of walking at various angles, but we only have this ground truth map. How might we use $G^{*}$ to generate training examples?

\subsection{Static Training Dataset}

We initially attempted to generate a static set of training examples. For each training example, we sample a region $(v_0, B)$ and a step count $n$, and initialize a search. We run $n$ steps of the search using an ``oracle'' decision function that uses $G^{*}$ to always make optimal decisions. The state of the search algorithm immediately preceding the $n$th step is the input for the training example, while the action taken by the oracle on the $n$th step is used to create a target output $O^{*}_{action} = \langle o^{*}_{walk}, o^{*}_{stop} \rangle$, $O^{*}_{angle} = \langle o^{*}_1, \ldots, o^{*}_a \rangle$. We can then train a CNN using gradient descent by back-propagating a cross entropy loss between $O_{action}$ and $O^{*}_{action}$, and, if $o^{*}_{walk} = 1$, a mean-squared error loss between $O_{angle}$ and $O^{*}_{angle}$.


However, we found that although the CNN can achieve high performance in terms of the loss function on the training examples, it performs poorly during inference. This is because $G$ is essentially perfect in every example that the CNN sees during training, as it is constructed by the oracle based on the ground truth map. During inference, however, the CNN may choose angles that are slightly off from the ones predicted by the oracle, resulting in small errors in $G$. Then, because the CNN has not been trained on imperfect inputs, these small errors lead to larger prediction errors, which in turn result in even larger errors.

\begin{figure}[t]
\begin{center}
	\includegraphics[width=0.9\linewidth]{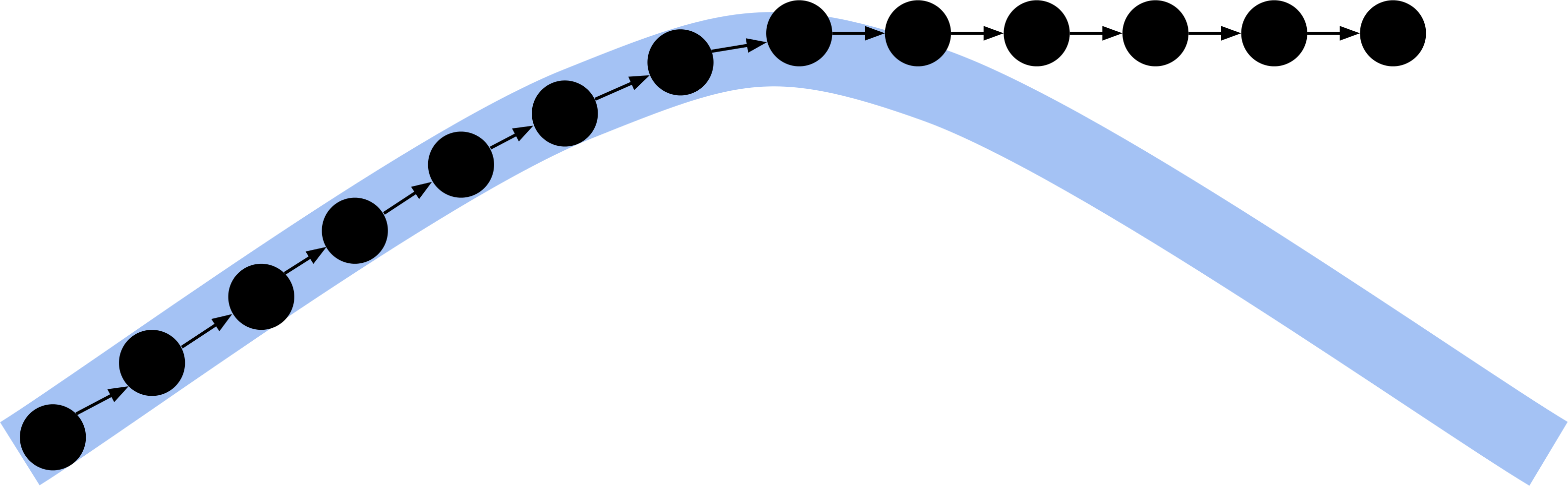}
\end{center}
	\caption{A CNN trained on static training examples exhibits problematic behavior during inference. Here, the system veers off of the road represented by the blue path.}
\label{fig:static_veer_off}
\vspace{-.1in}
\end{figure}

Figure \ref{fig:static_veer_off} shows a typical example of this snowball effect. The CNN does not output the ideal angle at the turn; this causes it to quickly veer off the actual road because it never saw such deviations from the road during training, and hence it cannot correct course.
We tried to mitigate this problem by using various methods to introduce noise on $G$ in the training examples. Although this reduces the scale of the problem, the CNN still yields low performance at inference time, because the noise that we introduce does not match the characteristics of the noise introduced inherently by the CNN during inference. Thus, we conclude a static training dataset is not suitable.

\subsection{Dynamic Labels}

We instead generate training examples dynamically by running the search algorithm with the CNN as the decision function \emph{during training}. As the CNN model evolves, we generate new training examples as well.



Given a region $(v_0, B)$, training begins by initializing an instance of the search algorithm $(G, S)$, where $G$ is the partial graph (initially containing only $v_0$) and $S$ is the vertex stack. On each training step, as during inference, we feed-forward the CNN to decide on an action based on the output layer, and update $G$ and $S$ based on that action.

In addition to deciding on the action, we also determine the action that an oracle would take, and train the CNN to learn that action. The key difference from the static dataset approach is that, here, $G$ and $S$ are updated based on the CNN output and not the oracle output; the oracle is only used to compute a label for back-propagation.

The basic strategy is similar to before. On each training step, based on $G^{*}$, we first identify the set of angles $R$ where there are unexplored roads from $S_{\text{top}}$. Next, we convert $R$ into a target output vector $O^{*}$. If $R$ is empty, then $o^{*}_{stop} = 1$. Otherwise, $o^{*}_{walk} = 1$, and for each angle $\theta \in R$, we set $o_i^{*} = 1$, where $i$ is the closest walkable angle to $\theta$. Lastly, we compute a loss between $O$ and $O^{*}$, and apply back-propagation to update the CNN parameters.



\begin{figure}[t]
\begin{center}
	\includegraphics[width=0.8\linewidth]{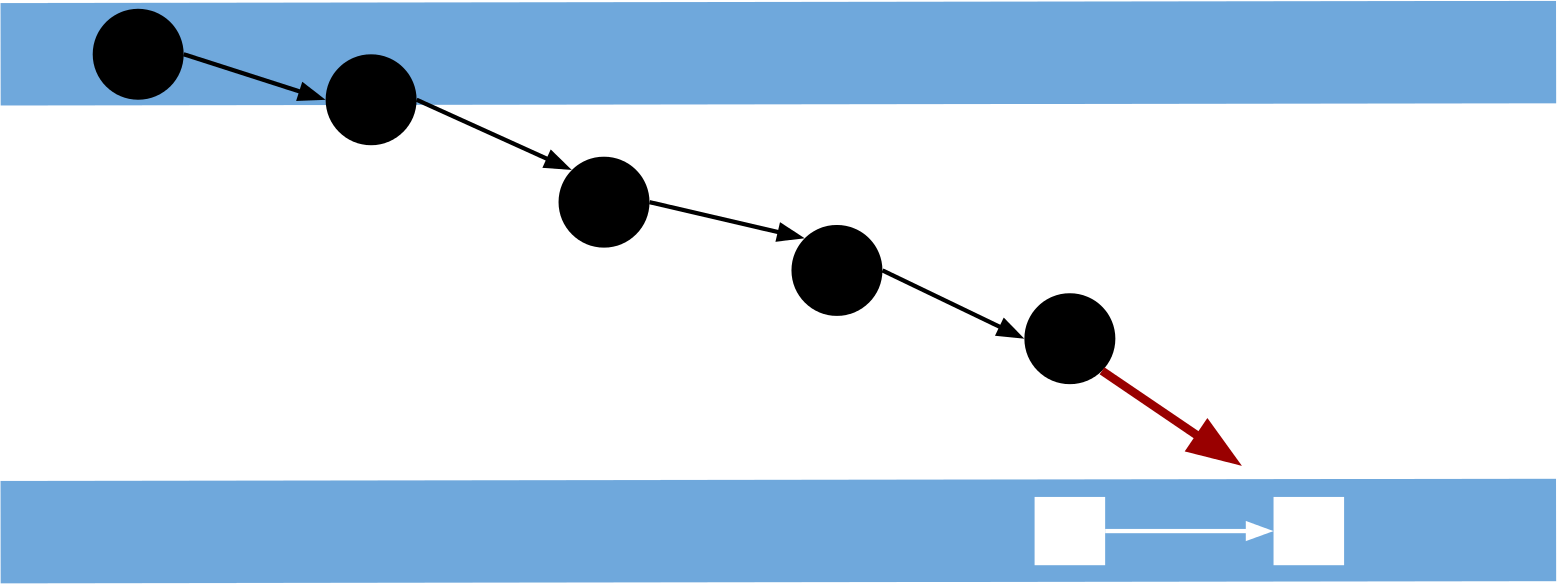}
\end{center}
	\caption{A naive oracle that simply matches $S_{\text{top}}$ to the closest location on $G^{*}$ fails, since it directs the system towards the bottom road instead of returning to the top road. Here, the black circles make up $G$, while the blue corresponds to the actual road position.}
\label{fig:naive_oracle}
\vspace{-.1in}
\end{figure}

A key challenge is how to decide where to start the walk in $G^*$  to pick the next vertex. The naive approach is to start the walk from the closest location in $G^{*}$ to $S_{\text{top}}$. However, as the example in Figure~\ref{fig:naive_oracle} illustrates, this approach can direct the system towards the wrong road when $G$ differs from $G^*$.

To solve this problem, we apply a {\it map-matching} algorithm to find a path in $G^{*}$ that is most similar to a path in $G$ ending at $S_{\text{top}}$. To obtain the path $p$ in $G$, we perform a random walk in $G$ starting from $S_{\text{top}}$. We stop the random walk when we have traversed a configurable number of vertices $w$ (we use $w = 10$), or when there are no vertices adjacent to the current vertex that haven't already been traversed earlier in the walk. Then, we match this path to the path $p^*$ in $G^*$ to which it is most similar. We use a standard map-matching method based on the Viterbi algorithm~\cite{vtrack}. If $v$ is the endpoint of the last edge in $p^{*}$, we start our walk in $G^{*}$ at $v$.

Finally, we maintain a set $E$ containing edges of $G^{*}$ that have already been explored during the walk. $E$ is initially empty. On each training step, after deriving $p^{*}$ from map-matching, we add each edge in $p^{*}$ to $E$. Then, when performing the walk in $G^{*}$, we avoid traversing edges that are in $E$ again.

\section{Evaluation} \label{sec:evaluation}


\noindent
\textbf{Dataset.} To  evaluate our approach, we assemble a large corpus of high-resolution satellite imagery and ground truth road network graphs covering the urban core of forty cities across six countries. For each city, our dataset covers a region of approximately 24 sq km around the city center. We obtain satellite imagery from Google at 60 cm/pixel resolution, and the road network from OSM (we exclude certain non-roads that appear in OSM such as pedestrian paths and parking lots). We convert the coordinate system of the road network so that the vertex spatial coordinate annotations correspond to pixels in the satellite images.

We split our dataset into a training set with 25 cities and a test set with 15 other cities. To our knowledge, we conduct the first evaluation of automatic mapping approaches where systems are trained and evaluated on entirely separate cities, and not merely different regions of one city, and also the first large-scale evaluation over aerial images from several cities. Because many properties of roads vary greatly from city to city, the ability of an automatic mapping approach to perform well even on cities that are not seen during training is crucial; the regions where automatic mapping holds the most potential are the regions where existing maps are non-existent or inaccurate.

\medskip
\noindent
\textbf{Baselines.} We compare RoadTracer with two baselines: DeepRoadMapper~\cite{mattyus2017deeproadmapper} and our own segmentation-based approach. Because the authors were unable to release their software to us, we implemented DeepRoadMapper, which trains a residual network with a soft intersection-over-union (IoU) loss function, extracts a graph using thresholding and thinning, and refines the graph with a set of heuristics and a missing connection classifier.

However, we find that the IoU loss results in many gaps in the segmentation output, yielding poor performance. Thus, we also implement our own segmentation approach that outperforms DeepRoadMapper on our dataset, where we train with cross entropy loss, and refine the graph using a four-stage purely heuristic cleaning process that prunes short segments, removes small connected components, extends dead-end segments, and merges nearby junctions.

\medskip
\noindent
\textbf{Metrics.} We evaluate RoadTracer and the segmentation schemes on TOPO~\cite{biagioni2012inferring}, SP~\cite{wegner2013higher}, and a new junction metric defined below. TOPO and SP are commonly used in the automatic road map inference literature~\cite{biagioni2012map, stanojevic2017kharita, wegner2015road, ahmed2015comparison}. TOPO simulates a car driving a certain distance from several seed locations, and compares the destinations that can be reached in $G$ with those that can be reached in $G^{*}$ in terms of precision and recall. SP generates a large number of origin-destination pairs, computes the shortest path between the origin and the destination in both $G$ and $G^{*}$ for each pair, and outputs the fraction of pairs where the shortest paths are similar (distances within 5\%).

However, we find that both TOPO and SP tend to assign higher scores to noisier maps, and thus don't correlate well with the usability of an inferred map. Additionally, the metrics make it difficult to reason about the cause of a low or high score.

Thus, we propose a new evaluation metric with two goals: (a) to give a score that is representative of the inferred map's practical usability, and (b) to be interpretable. Our metric compares the ground truth and inferred maps junction-by-junction, where a junction is any vertex with three or more edges. We first identify pairs of corresponding junctions $(v, u)$, where $v$ is in the ground truth map and $u$ is in the inferred map. Then, $f_{v,\text{correct}}$ is the fraction of incident edges of $v$ that are captured around $u$, and $f_{u,\text{error}}$ is the fraction of incident edges of $u$ that appear around $v$. For each unpaired ground truth junction $v$, $f_{v,\text{correct}} = 0$, and for each unpaired inferred map junction $u$, $f_{u,\text{error}} = 1$. Finally, if $n_{\text{correct}} = \sum_v f_{v,\text{correct}}$ and $n_{\text{error}} = \sum_u f_{u,\text{error}}$, we report the correct junction fraction $F_{\text{correct}} = \frac{n_{\text{correct}}}{\text{\# junctions in } G^*}$ and error rate $F_{\text{error}} = \frac{n_{\text{error}}}{n_{\text{error}} + n_{\text{correct}}}$.

TOPO and our junction metric yield a precision-recall curve, while SP produces a single similar path count.

\begin{figure}[t]
\begin{center}
	\includegraphics[width=\linewidth]{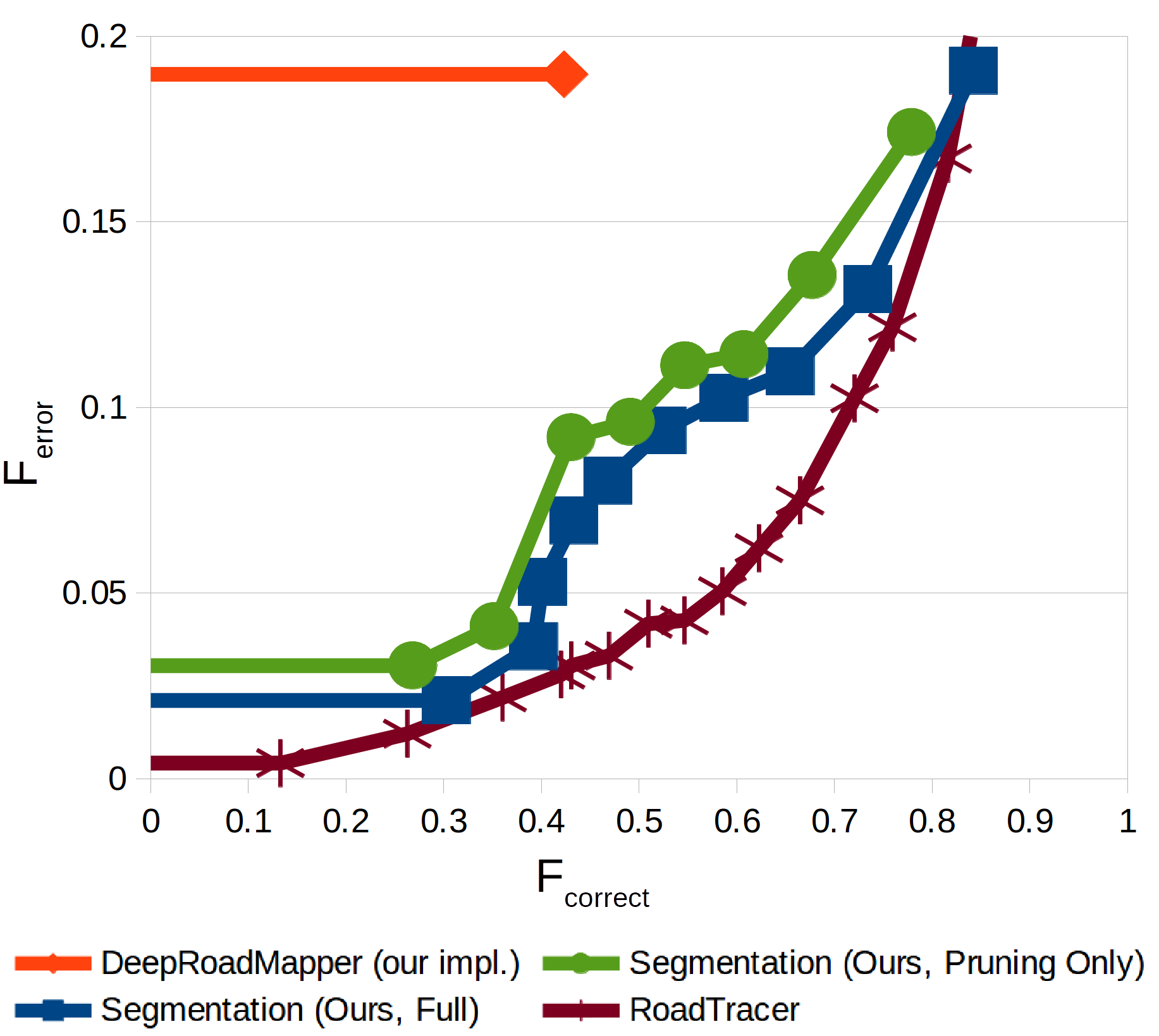}
\end{center}
	\caption{Average $F_{\text{correct}}$ and $F_{\text{error}}$ over the 15 test cities.}
\label{fig:result_intersections}
\end{figure}

\begin{figure}[t]
\begin{center}
	\includegraphics[width=\linewidth]{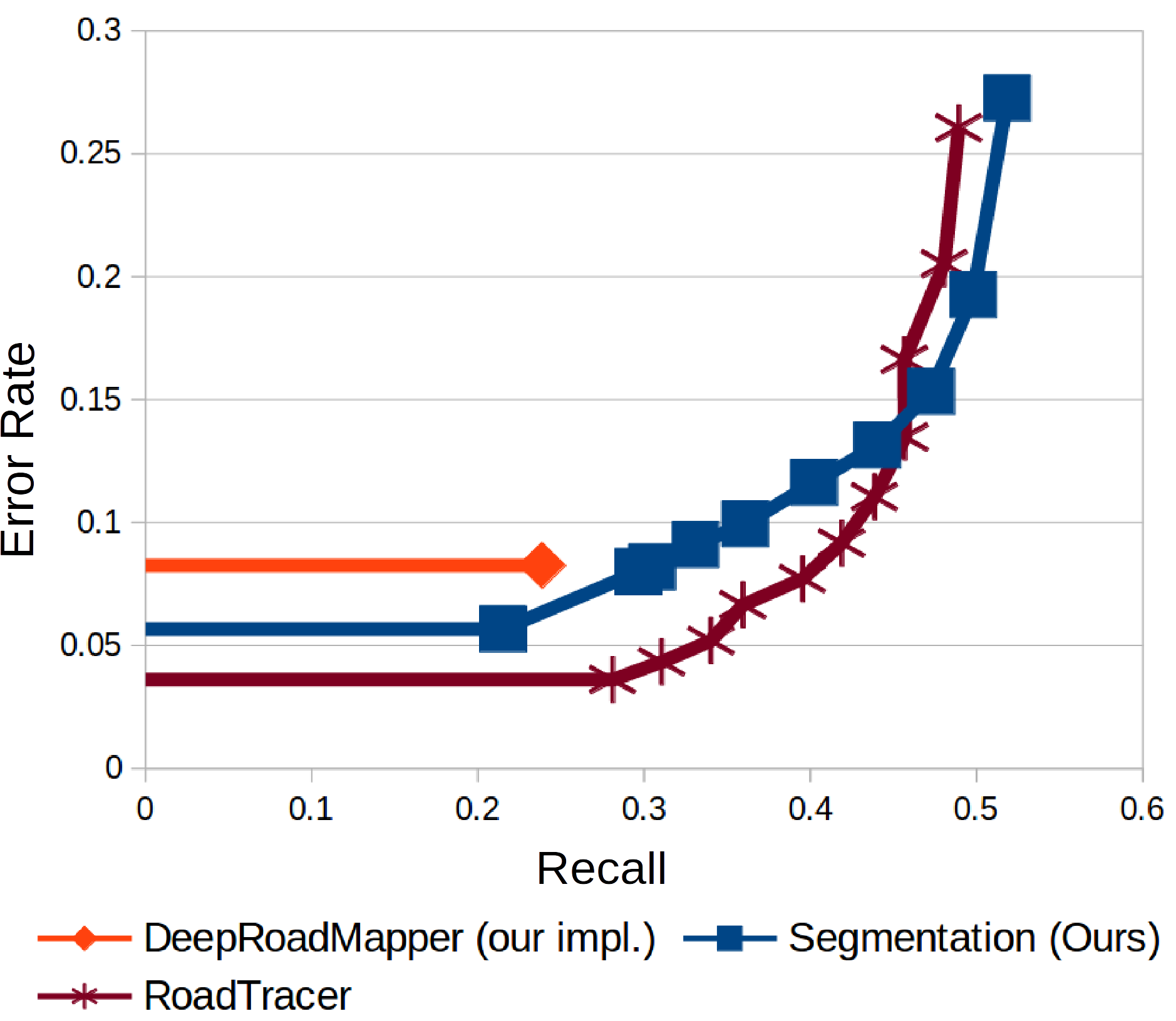}
\end{center}
	\caption{Average TOPO recall and error rate over the test cities.}
\label{fig:result_topo}
\end{figure}

\begin{table}[t]
    \centering
    \begin{tabular}{| c | c | c | c | c |}
        \hline
        Scheme & Correct & Long & Short & NoPath \\
        \hline
        DeepRoadMapper & 0.21 & 0.29 & 0.03 & 0.47 \\
        \hline
        Seg. (Ours) & 0.58 & 0.14 & 0.27 & 0.01 \\
        \hline
        RoadTracer & \textbf{0.72} & 0.16 & 0.10 & 0.02 \\
        \hline
    \end{tabular}
    \vspace*{.1in}
    \caption{SP performance. For each scheme, we only report results for the threshold that yields the highest correct shortest paths. Long, Short, and NoPath specify different reasons for an inferred shortest path being incorrect (too long, too short, and disconnected).}
    \label{tab:sp}
\end{table}

\begin{figure}[t]
\begin{center}
	\includegraphics[width=\linewidth]{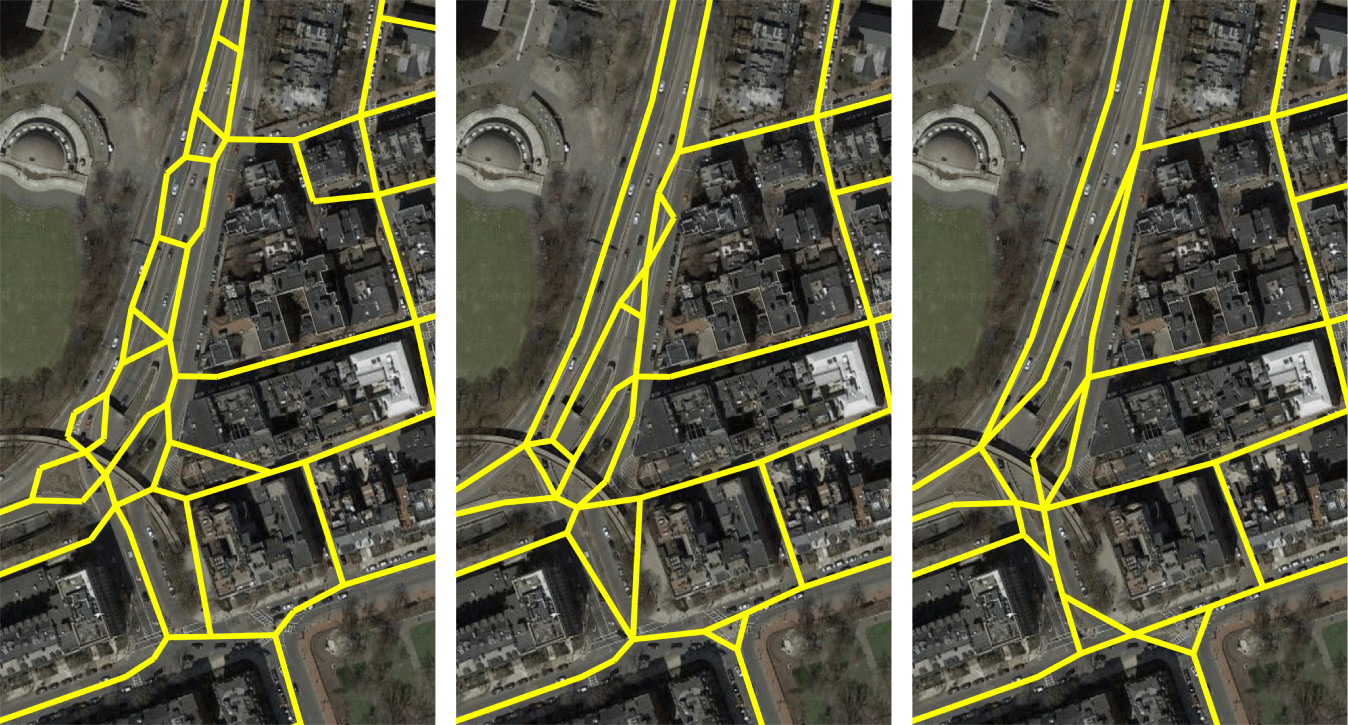}
\end{center}
	\caption{Tradeoff between error rate and recall in a small crop from Boston as we increase the threshold for our segmentation approach. The junction metric error rates in the crop from left to right are 18\%, 13\%, and 8\%. The map with 18\% error is too noisy to be useful.}
\label{fig:tradeoff}
\end{figure}

\begin{figure*}
\begin{center}
	\includegraphics[width=\linewidth]{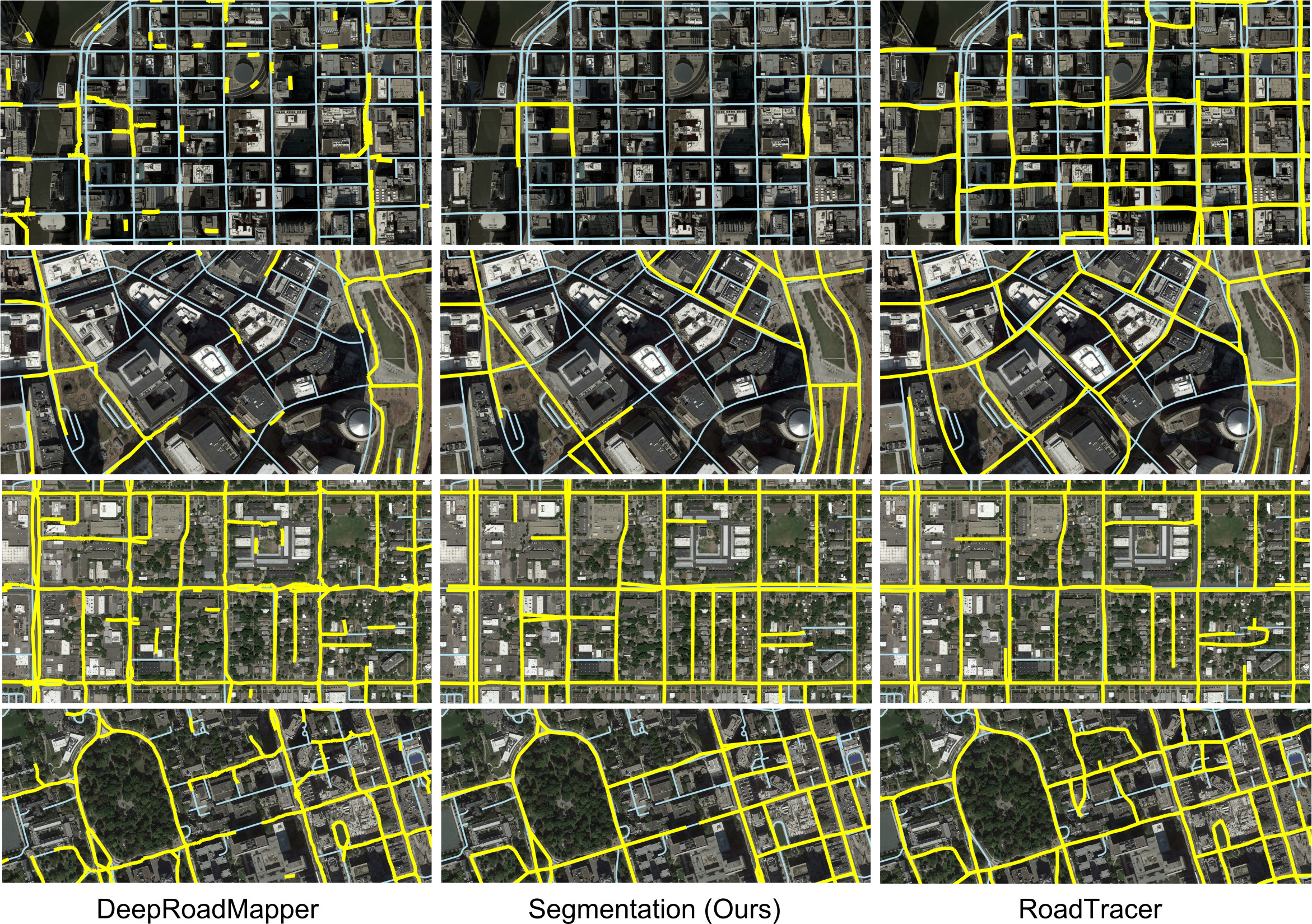}
\end{center}
	\caption{Comparison of inferred road networks in Chicago (top), Boston, Salt Lake City, and Toronto (bottom). We overlay the inferred graph (yellow) over ground truth from OSM (blue). Inferred graphs correspond to thresholds that yield 5\% average $F_{\text{error}}$ for RoadTracer and our segmentation approach, and 19\% $F_{\text{error}}$ for DeepRoadMapper (as it does not produce results with lower average error).}
\label{fig:qualitative}
\end{figure*}

\medskip
\noindent
\textbf{Quantitative Results.} We evaluate performance of the three methods on 15 cities in the test set. We supply starting locations for RoadTracer by identifying peaks in the output of our segmentation-based approach. All three approaches are fully automated.

Both RoadTracer and the segmentation approaches have parameters that offer a tradeoff between recall and error rate ($1-\text{precision}$). We vary these parameters and plot results for our junction metric and TOPO on a scatterplot where one axis corresponds to recall and the other corresponds to error rate. For DeepRoadMapper and our segmentation approach, we vary the threshold used to produce a binary mask. We find that the threshold does not impact the graph produced by DeepRoadMapper, as the IoU loss pushes most outputs to the extremes, and thus only plot one point. For RoadTracer, we vary the walk-stop action threshold $T$.

We report performance in terms of average $F_{\text{correct}}$ and $F_{\text{error}}$ across the test cities in Figure \ref{fig:result_intersections}, and in terms of average TOPO precision and recall in Figure \ref{fig:result_topo}.

On the junction metric, RoadTracer has a better $F_{\text{error}}$ for a given $F_{\text{correct}}$. The performance improvement is most significant when error rates are between 5\% and 10\%, which is the range that offers the best tradeoff between recall and error rate for most applications---when error rates are over 10\%, the amount of noise is too high for the map to be usable, and when error rates are less than 5\%, too few roads are recovered (see Figure \ref{fig:tradeoff}). When the error rate is 5\%, the maps inferred by RoadTracer have 45\% better average recall ($F_{\text{correct}}$) than those inferred by the segmentation approach (0.58 vs 0.40).

On TOPO, RoadTracer has a lower error rate than the segmentation approaches when the recall is less than 0.43. Above 0.43 recall, where the curves cross, further lowering $T$ in RoadTracer yields only a marginal improvement in recall, but a significant increase in the error rate. However, the segmentation approach outperforms RoadTracer only for error rates larger than 0.14; we show in Figure \ref{fig:tradeoff} that inferred maps with such high error rates are not usable.

We report SP results for the thresholds that yield highest number of correct shortest paths in Table \ref{tab:sp}. RoadTracer outperforms the segmentation approach because noise in the output of the segmentation approach causes many instances where the shortest path in the inferred graph is much shorter than the path in the ground truth graph.


Our DeepRoadMapper implementation performs poorly on our dataset. We believe that the soft IoU loss is not well-suited to the frequency of occlusion and complex topology found in the city regions in our dataset.

\bigskip
\noindent
\textbf{Qualitative Results.} In Figure \ref{fig:qualitative}, we show qualitative results in crops from four cities from the test set: Chicago, Boston, Salt Lake City, and Toronto. For RoadTracer and our segmentation approach, we show inferred maps for the threshold that yields 5\% average $F_{\text{error}}$. DeepRoadMapper only produces one map.

RoadTracer performs much better on frequent occlusion by buildings and shadows in the Chicago and Boston regions. Although the segmentation approach is able to achieve similar recall in Boston on the lowest threshold (not shown), several incorrect segments are added to the map. In the Salt Lake City and Toronto regions, performance is comparable. DeepRoadMapper's soft IoU loss introduces many disconnections in all four regions, and the missing connection classifier in the post-processing stage can only correct some of these.

We include more outputs in the supplementary material, and make our code, full-resolution outputs, and videos showing RoadTracer in action available at \url{https://roadmaps.csail.mit.edu/roadtracer}.

\section{Conclusion}

On the face of it, using deep learning to infer a road network graph seems straightforward: train a CNN to recognize which pixels belong to a road, produce the polylines, and then connect them. But occlusions and lighting conditions pose challenges, and such a segmentation-based approach requires complex post-processing heuristics. By contrast, our iterative graph construction method uses a CNN-guided search to directly output a graph. We showed how to construct training examples dynamically for this method, and evaluated it on 15 cities, having trained on aerial imagery from 25 entirely different cities. To our knowledge, this is the largest map-inference evaluation to date, and the first that fully separates the training and test cities. Our principal experimental finding is that, at a 5\% error rate, RoadTracer correctly captures 45\% more junctions than our segmentation approach (0.58 vs 0.40). Hence, we believe that our work presents an important step forward in fully automating map construction from aerial images.

\section{Acknowledgements}

This research was supported in part by the Qatar Computing Research Institute (QCRI).


{\small
\bibliographystyle{ieee}
\bibliography{egbib}
}

\end{document}